\documentclass[10pt,leqno]{amsart}
\usepackage{graphicx}
\usepackage{indentfirst,csquotes}

\usepackage{lipsum}
\usepackage{amssymb,amsthm,amsmath}
\usepackage{xcolor,paralist,hyperref,fancyhdr,etoolbox}

\usepackage{times}
\usepackage{amsmath}
\usepackage[table]{xcolor}
\usepackage{float}
\usepackage[most]{tcolorbox}
\usepackage{xcolor}
\usepackage{lipsum}
\usepackage{microtype}
\usepackage{amssymb}
\usepackage{graphicx}
\usepackage[most]{tcolorbox}
\usepackage{booktabs}
\usepackage{url}
\usepackage{multirow}

\hypersetup{ colorlinks=true, linkcolor=black, filecolor=black, urlcolor=black }

\usepackage{lipsum}
\usepackage{hyperref}

\hypersetup{
  pdftitle={Query-Focused Event Summarization: A Dataset and Benchmark},
  pdfauthor={Chenyu Hu and Bang Wang},
  pdfkeywords={query-focused summarization, event summarization, dataset, retrieval}
}

\begin{document}

\title{Query-Focused Event Summarization: A Dataset and Benchmark}

\author[Chenyu Hu]{Chenyu Hu}
\address{School of Electronic Information and Communications, Huazhong University of Science and Technology}
\email{huchenyu@hust.edu.cn}

\author[Bang Wang]{Bang Wang\textsuperscript{*}}
\address{School of Electronic Information and Communications, Huazhong University of Science and Technology}
\email{wangbang@hust.edu.cn}

\thanks{\textsuperscript{*}Corresponding author.}

\date{\today}

\maketitle

\let\thefootnote\relax

\begin{abstract}
\textit{Thematic corpus} refers to a corpus of semantically coherent documents that collectively describe different aspects of a shared thematic event. Thematic corpus typically contains hundreds or even thousands of documents. While users' concerns on the thematic event often span multiple dimensions, Query-Focused Summarization (QFS) aims to generate summaries tailored to users' queries. However, existing QFS datasets lack event summarization, and most QFS methods struggle with large-scale datasets. To address these challenges, we propose \textbf{QFES} (Query-Focused Event Summarization) task and construct \textbf{QFESum} dataset, which contains 8 thematic events, 16{,}684 documents, and 104 queries. Furthermore, we introduce a two-stage QFES framework consists of Query-Focused Retrieval with Adaptive Thresholding (RAT) and Query-Focused Summarization based on Hierarchical Clustering (SHC). Experimental results on QFESum show that RAT and SHC perform consistently better than baselines, demonstrating their effectiveness in QFES. The dataset and code are publicly available at \url{https://github.com/sarcasm-hcy02/QFES-QFESum}.
\end{abstract} 

\section{Introduction}
A thematic corpus refers to a collection of semantically coherent documents or reports that collectively describe different aspects of a shared thematic event. Thematic events such as \textit{``global financial crisis''} span multiple dimensions including economy, society, and politics, and involve massive heterogeneous thematic corpus. Fully understanding such thematic events requires substantial human effort, making automatic summarization highly valuable in real-world thematic events analysis. As a result, summarization techniques have been widely explored: Multi-Documents Event Summarization(MDES)~\cite{pratapa2025scaling} aims to generate a coherent and concise event summary that captures the main content from documents. However, MDES produces only a generic summary for the entire thematic corpus, whereas different users often have distinct interests. For a given thematic event, different people may focus on different aspects: environmental organizations emphasize ecological impacts, investors pay attention to financial risks, while policymakers are concerned with regulatory responsibilities.


Therefore, Query-Focused Summarization (QFS) has been proposed to generate summaries tailored to users’ distinct queries.
However, existing QFS datasets and methods lack designs specifically tailored to events, while event summarization is one of the most widely demanded and frequently used summarization scenarios. Thus we propose Query-Focused Event Summarization task (QFES), which aims to generate event summaries focused on users' query from a thematic corpus. 

Compared with QFS, QFES differs substantially in terms of datasets, algorithms, and evaluation. At the dataset level, a QFES dataset requires thematic corpus, together with corresponding queries, where each query represents a specific aspect of the thematic event. Moreover, each document in the corpus should be associated with query labels indicating which aspects the document is related to. At the algorithmic level, a QFES method is expected to first retrieve documents relevant to a given query from the thematic corpus, and then generate a summary whose content is focused on the query-specified aspect of the thematic event. At the evaluation level, in addition to conventional word-level metrics such as ROUGE, QFES summaries should be assessed crucially from an event-granularity perspective. That means evaluation should examine whether the events in the generated summary can be properly semantic matched with those in the reference summary.

Despite existing progress in QFS and MDES research, several limitations remain when applied to QFES:

1) QFS datasets lack event summary. Popular QFS datasets such as SQuALITY~\cite{wang2022squality} and QMSum~\cite{zhong2021qmsum} primarily focus on short stories or domain-specific meeting records, whose query designs are not aligned with events.
2) MDES datasets are either small in scale or lack query. For example, DUC~\cite{dang2005overview} is limited in size and not open-source,  NEWTS~\cite{bahrainian2022newts} only supports multi-perspective summarization of single-news input. Although T17~\cite{ghalandari2020examining} and CRISIS~\cite{tran2015timeline} contain a set of thematic corpus, each only with a timeline event summary, they do not provide queries.
3) Most QFS methods exhibit significant performance degradation on large-scale datasets. For instance, SEGENC~\cite{vig2022exploring} relies on Longformer-Encoder-Decoder LED~\cite{beltagy2020longformer}, which struggles to maintain global consistency and cross-document relations when scaling to large-scale datasets that contains hundreds or thousands of documents.
4) Existing QFS methods do not quantify the impact of retrieval quality on summarization performance. Documents retrieval is a crucial prerequisite of QFES. However, existing QFS studies~\cite{wang2022squality} typically assume retrieval to be perfect, lacking systematic evaluation of retrieval.

To address the above limitations, we construct a large-scale query-focused event summarization dataset named \textbf{QFESum}, which is built based on T17 and CRISIS through an annotation framework in which annotators label the key core steps and LLMs are used for auxiliary annotation. QFESum augments existing thematic corpus with multiple queries and document-level relevance annotations. The construction process consists of three stages:
(1) Query Initialization, where a set of initial queries is obtained from the thematic corpus;
(2) Query-Focused Summary Construction, where query-focused reference summaries are manually selected and the queries are manually refined;
(3) Query-Related Document Annotation, where documents are annotated with relevance labels for each query.

In addition, we introduce a two-stage summarization framework for QFES, consisting of Query-Focused Retrieval with Adaptive Thresholding (RAT) and Query-Focused Summarization based on Hierarchical Clustering (SHC). 

For retrieving documents relevant to a given query, RAT first samples a subset of thematic corpus, then generates a query-specific dynamic threshold, and finally determines document relevance based on this threshold or resorts to an additional LLM verification when necessary. 

Subsequently, SHC generates query-focused summaries over the retrieved documents, consists of three stages:
(1) Event Extraction, where a query-focused set of events is extracted from the retrieved documents;
(2) Key Events Selection, where key events are selected from the event set via a hierarchical selection mechanism;
(3) Coreference Resolution, where coreferential events among the selected key events are resolved to form the final summary.  Experimental results on QFESum show that both RAT and SHC consistently outperform all baselines, demonstrating their effectiveness in handling complex thematic corpus.

\section{QFES Task and QFESum Dataset}
The QFES task is defined as follows: Given a query $q$ and a thematic corpus $\mathcal{D}$, the QFES objective is to first retrieve query-related documents from the thematic corpus and then generate a query-focused event summary. Since thematic corpora often contain a large number of documents, the retrieval step is usually necessary to reduce the computational burden of subsequent event summarization and mitigate the interference of irrelevant documents.

\par
Conducting experiments on the QFES task requires generating multiple queries for each thematic corpus, preparing query-focused reference summaries for these queries, and annotating the query relevance of each document in the thematic corpus. That is, we need a query set $\mathcal{Q}=\{q_i\}$, query-related document set $\mathcal{D}_i^\text{r} \subseteq \mathcal{D}$ each only containing the documents related to the query $q_i$, and the query-focused reference summaries $\{S_i\}$. We note that although existing datasets for event timeline summarization, such as the \texttt{T17}~\cite{ghalandari2020examining} and \texttt{CRISIS}~\cite{tran2015timeline}, cannot be directly used for the QFES task due to lack of $\{q_i\}, \{\mathcal{D}_i^\text{r}\}, \{S_i^\text{r}\}$, they have provided a corpus $\mathcal{D}$ associated with timeline summaries $\{S_\text{tls}\}$.

\par
\begin{figure*}[t]
    \centering
    \includegraphics[width=\textwidth]{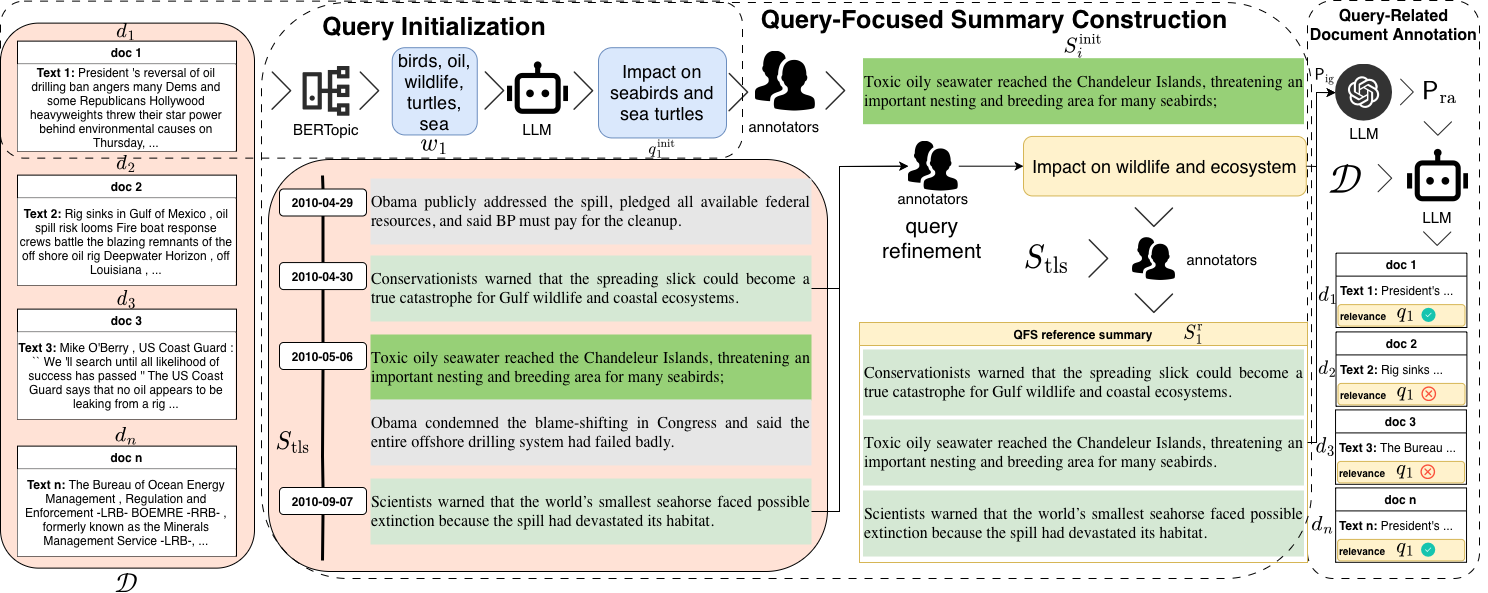}
    \caption{The overall construction of QFESum. In $S_{\text{tls}}$, the dark green events are relevant to $q_1^{\text{init}}$, while the light green events are relevant to the thematic event $\mathcal{E}_t$. The input consists of the content shown in the red boxes, while the output corresponds to the content shown in the yellow boxes.}
    \label{pic:QFESum}
\end{figure*}

In this paper, we construct the QFESum dataset for the QFES task based on the \texttt{T17} and \texttt{CRISIS} dataset, the construction process is shown in Figure~\ref{pic:QFESum}. Notice that both datasets contain multiple thematic corpora, each of which consists of distinct documents and timeline summaries. In addition, each corpus is associated with a thematic label. We use the corpora from both datasets and process them on a per-corpus basis. For a thematic corpus $\mathcal{D}$, its timeline summary comprises multiple chronologically timestamped entries, where each entry contains one or more sentences describing representative events occurring at the corresponding timestamp. Two graduate annotators were invited to construct the \textit{reference event set} $\mathcal{E}$. Specifically, they concatenated all event sentences from all timestamps in chronological order and removed duplicate events appearing under the same timestamp. They then cross-checked whether the constructed $\mathcal{E}$ was identical across annotators; in cases of disagreement, the third annotator was responsible for determining the final version. 

\par
For a given $\mathcal{D}$ associated with $\mathcal{E}$, we construct a query set $\{q_i\}$, a relevant document set $\mathcal{D}_i^\text{r}$ and a reference summary $S_i^\text{r}$ for each query $q_i$ to form the QFESum dataset. For example, a corpus in \texttt{T17} is tagged with the \texttt{BPoil} label, which contains 1307 documents. Its reference event set $\mathcal{E}$ contains 363 events with in total 7035 words. We construct a query set $\mathcal{Q}$ contains 21 queries, and one query $q_i$ is "\texttt{impact on wildlife and ecosystems}". For this query, its  associated relevant document set $\mathcal{D}_i^\text{r}$ contains 189 documents, and the corresponding reference summary $S_i^\text{r}$ contains 21 events and 393 words. In the following, we describe the detailed process of constructing the query set ${q_i}$, the reference summary $S_i^\text{r}$, and the relevant document set $\mathcal{D}_i^\text{r}$.

\paragraph{Query Initialization:} For each $\mathcal{D}$, we first apply the BERTopic clustering~\cite{grootendorst2022bertopic} to produce a topic set $\mathcal{T}=\{\tau_k \}$ ordered by their topic mention frequencies. Each topic $\tau_k$ is represented by an associated keyword set, from which we select the top-K highest-ranked keywords, denoted as $\{w_k\}$, to represent the topic $\tau_k$. For each topic $\tau_k$, we design a \textit{query generation}  prompt $\mathsf{P}_\text{qg}$ as sketched in Appendix~\ref{Fig:QueryGenerationPrompt}, which is input into the ChatGPT-4o~\cite{roumeliotis2023chatgpt} for transforming the selected keywords $\{w_k\}$ into a readable phrase as the initial query $q_i^\text{init}$. Let $\mathcal{Q}^\text{init}=\{q_i^\text{init}\}$ denote the set of initial queries.

\paragraph{Query-Focused Summary Construction:} We employ the two aforementioned graduate annotators to construct a query-focused summary $S_i^\text{init}$ for each $q_i^\text{init}$ by manually examining all sentences in $\mathcal{E}$. If a sentence containing event(s) is considered by the annotators as relevant to the query, it is included into the summary, yielding an initial query and summary pair $(q_i^\text{init}, S_i^\text{init})$ and $S_i^\text{init} \subseteq \mathcal{E}$. During this process, annotators have observed that some summary contains too few events and some has overmuch semantic repetitions with others. To address these, annotators further improves the initial queries in $\mathcal{Q}^\text{init}$ to align their semantics with those relevant events in $\mathcal{E}$.

\par
To reduce semantic redundancy and ensure comprehensive event coverage, annotators conduct a systematic review on all sentences in $\mathcal{E}$ to evaluate its thematic coverage and event distinctiveness, manually obtaining a set of distinctive event themes. By examining event themes, annotators select queries from $\mathcal{Q}^\text{init}$ to ensure that those selected queries collectively cover all event themes in $\mathcal{E}_t$, yet without too much semantic overlaps. Let $\mathcal{Q}^\text{cand}$ denote the set of selected queries.

\par
For each $q_i^\text{cand} \in \mathcal{Q}^\text{cand}$, annotators reexamine its associated summary $S_i^\text{init}$. If the events in $S_i^\text{init}$ correspond to more than five distinct timestamps in $\mathcal{S}_\text{tls}$, then its query and its associated summary is kept into as the final query $q_i$; Otherwise, annotators refine the query $q_i^\text{cand}$ by manually changing its expression but without changing its event theme, typically by increasing the level of abstraction and generality of $q_i^\text{cand}$. For example, \textit{impact on seabirds and sea turtles''} can be revised as \textit{impact on wildlife and ecosystem''}. The query refinement is associated with a manually re-extracted summary, until the summary events cover more than five timestamps.

\par
After query refinement, we obtain the final query set $\mathcal{Q}={q_i}$. For each query $q_i$, the three annotators independently selected the corresponding event summaries from the thematic event set. We then compared the event summaries selected by the three annotators and observed an average agreement rate of 75\%. In cases of disagreement, the third annotator made the final decision, resulting in the final query-summary pair, denoted as $(q_i, S_i^\text{r})$.

\begin{table*}[!t]
\centering
\small
\resizebox{\textwidth}{!}{
\begin{tabular}{lcccccccc}
\toprule
\textbf{Corpus} &
\textbf{BPoil} &
\textbf{Finan} &
\textbf{Iraq} &
\textbf{Syria\_T17} &
\textbf{Egypt} &
\textbf{Libya} &
\textbf{Syria\_Crisis} &
\textbf{Yemen} \\
\midrule

\textbf{documents\_num\_in\_$\mathcal{D}$} & 1307 & 256 & 315 & 624 & 2126 & 2840 & 5170 & 4046 \\
\textbf{word\_num\_in\_$S_{\text{tls}}$} & 7035 & 8110 & 7705 & 2893 & 3803 & 7003 & 3832 & 3236 \\
\textbf{event\_num\_in\_$S_{\text{tls}}$} & 552 & 601 & 559 & 312 & 358 & 635 & 318 & 341 \\
\textbf{queries\_num\_in\_$\mathcal{Q}$} & 21 & 21 & 14 & 8 & 10 & 14 & 8 & 8 \\

\textbf{average\_documents\_num\_in\_$\mathcal{D}^{r}_{i}$} & 222.57(17.03\%) & 75.19(29.37\%) & 82.07(26.05\%) & 286.12(45.85\%) & 786.70(37.00\%) & 599.29(21.10\%) & 1146.22(22.17\%) & 899.38(22.23\%) \\
\textbf{max\_documents\_num\_in\_$\mathcal{D}^{r}_{i}$} & 691(52.87\%) & 204(79.69\%) & 175(55.56\%) & 416(66.67\%) & 1504(70.74\%) & 1215(42.78\%) & 2086(40.35\%) & 1613(39.87\%) \\
\textbf{min\_documents\_num\_in\_$\mathcal{D}^{r}_{i}$} & 19(1.45\%) & 16(6.25\%) & 7(2.22\%) & 8(1.28\%) & 10(0.47\%) & 14(0.49\%) & 8(0.15\%) & 8(0.20\%) \\

\textbf{average\_event\_num\_in\_$S^{r}_{i}$} & 21.43 & 28.43 & 16.00 & 30.12 & 18.40 & 17.36 & 21.62 & 18.88 \\
\textbf{max\_event\_num\_in\_$S^{r}_{i}$} & 47 & 63 & 35 & 91 & 37 & 47 & 43 & 30 \\
\textbf{min\_event\_num\_in\_$S^{r}_{i}$} & 7 & 6 & 5 & 9 & 5 & 5 & 6 & 8 \\

\bottomrule
\end{tabular}
}
\caption{Some statistics of QFESum across the eight corpus.}
\label{tab:QFESumStatistics}
\end{table*}

\par
\paragraph{Query-Related Document Annotation:} For each query $q_i \in \mathcal{Q}$, we annotate its related documents $\mathcal{D}_i \subseteq \mathcal{D}$ as follows. A document $d_j \in \mathcal{D}$ is labeled as relevant to $q_i$, if there exists at least one sentence in this document relevant to $q_i$. Specifically, we first design an \textit{instruction generation} prompt $\mathsf{P}_{\text{ig}}$ including a query and its summary as shown in Figure~\ref{fig:instruction generation prompt}, which is used to command the ChatGPT-4o to output another query-specific \textit{relevance annotation} prompt $\mathsf{P}_{\text{ra}}$, an example of $\mathsf{P}_{\text{ra}}$ is shown in Figure~\ref{pic:prompt_annatation}.

\par
We segment a document $d_j$ into a set of sentences $\{s_k\}$.A sentence $s_k$ represents an event. Based on $\mathsf{P}_\text{ra}$, we input $s_k$ and $q_i$ into the Qwen2.5-7B~\cite{yang2025qwen3} to determine whether $s_k$ is related to $q_i$. If any sentence $s_k \in d_j$ is determined as relevant to the query $q_i$, then the document $d_j$ is annotated as relevant to the query. Finally, the three aforementioned annotators independently conducted relevance verification on randomly sampled 10\% of the labeled results. The results show that the prompt-driven relevance annotations achieved an average accuracy of 97\% compared with manual labeling across the samples checked by the three annotators.

\par
After the annotation and verification, each corpus $\mathcal{D}$ is associated with a query set $\{q_i\}$, and each query is associated with a relevant document set $\{\mathcal{D}_i^\text{r}\}$ as well as the query-focused summary $S_i^\text{r}$. Finally, the QFESum dataset contains eight of such corpus, four from the \texttt{T17} and four from the \texttt{CRISIS} dataset. More statistics are presented in Table~\ref{tab:QFESumStatistics}.

\begin{figure}[t]
    \centering
    \includegraphics[width=\columnwidth]{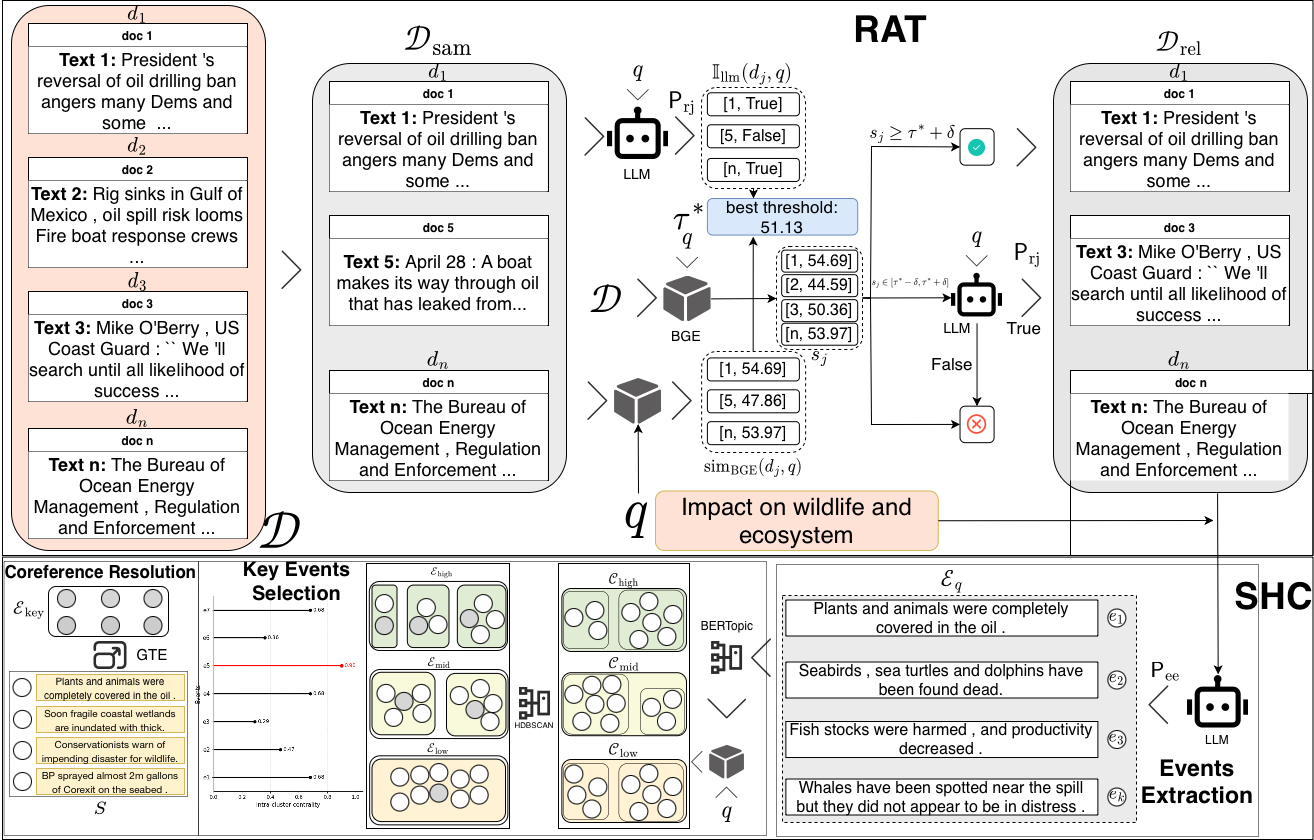}
    \caption{The overall architecture of RAT and SHC. The input consists of the content shown in the red boxes($\mathcal{D}$ and $q$), while the output corresponds to the content shown in the yellow boxes($S$).}
    \label{pic:RatShc}
\end{figure}

\section{Query-Focused Event Summarization}
We propose a framework to experiment the constructed QFESum dataset for the QFES task,as shown in Figure~\ref{pic:RatShc} which consists of two stages: Query-Focused Retrieval with Adaptive Thresholding (RAT) and Query-Focused Summarization based on Hierarchical Clustering (SHC).

\subsection{Query-Focused Retrieval with Adaptive Thresholding}\label{sec:RAT}
Given the input query $q$ and corpus $\mathcal{D}$, the RAT module is to retrieve relevant documents $\{d_j\} \subseteq \mathcal{D}$ via a hybrid approach with dense retrieval first and LLM filtering next. We use the BGE~\cite{chen2024bge} as the dense retrieval for primary screening, which retrieves documents according to the semantic similarity score $\mathrm{sim}_\text{BGE}(d_j, q)$ between $d_j$ and $q$. However, it is not optimal for setting a fixed similarity threshold for different queries, as the semantic distributions vary across queries. To address this issue, we design an adaptive thresholding mechanism as follows.

\par
At first, we randomly sample 10\% documents from $\mathcal{D}$ to form a subset $\mathcal{D}_\text{sam}$. For each $d_j \in \mathcal{D}_\text{sam}$, we compute $s_j = \mathrm{sim}_\text{BGE}(d_{j}, q)$. We also design a \textit{relevance judgement} prompt $\mathsf{P}_\text{rj}$ as shown in Figure~\ref{fig:prompt_relevance_judgment} including $d_j$ and $q$, and leverage an LLM (Qwen2.5-7B) to output its binary relevance judgement, denoted by $\mathbb{I}_\text{llm}(d_j, q)$. For each $d_j \in \mathcal{D}_\text{sam}$, we construct a set $\mathcal{D}_j = \{d_h\}$ with $d_h \in \mathcal{D}_\text{sam}$ and $s_h \geq s_j$. We compute the accuracy of $\mathcal{D}_j$ for setting $s_j$ as the retrieval threshold by
\[
    \mathrm{Acc}(s_j)= \frac{1}{|\mathcal{D}_j|}                      \sum_{h=1}^{|\mathcal{D}_j|} \mathbb{I}_\text{llm}(d_h, q).
\]
Among all $\{s_j\}$, we select the one with the largest $\mathrm{Acc}(s_j)$ as the retrieval threshold, denoted by $\tau^*$.

\par
After obtaining $\tau^*$, we compute the BGE similarities for all documents in $\mathcal{D}$. For a document $d_j$ with similarity $s_j$, if $s_j \geq \tau^* + \delta$ ($\delta=0.03$ in experiments), then this document is regarded as a relevant one. If $s_j \in [\tau^* - \delta, \tau^* + \delta]$, we further use the prompt $\mathsf{P}_\text{rj}$ its relevance judgement by an LLM (Qwen2.5-7B). If $\mathbb{I}_\text{llm}(d_j, q)=1$, then this document is also regarded as a relevant one. Finally, we obtain the relevant document set $\mathcal{D}_\text{rel}$.

\subsection{Query-Focused Summarization based on Hierarchical Clustering}\label{sec:SHC}
Given the query $q$ and its relevant documents $\mathcal{D}_\text{rel}$, the SHC is to generate an event summary $S$ as follows: It first extracts all events $\mathcal{E}_q$ from $\mathcal{D}_\text{rel}$, and then selects key events with coreference resolution for producing the summary.

\paragraph{Event Extraction:} We design a query-focused \textit{event extraction} prompt $\mathsf{P}_\text{ee}$ including $q$ and $\mathcal{D}_\text{rel}$, which is then input into an LLM (Qwen2.5-7B) to obtain $\mathcal{E}_q = \{e_k \}$.

\paragraph{Key Events Selection:} We use the BGE model to encode $\mathcal{E}_q$ into $\{
\mathbf{e}_k\}$ that are then input in the BERTopic clustering~\cite{grootendorst2022bertopic} to form event clusters $\mathcal{C}=\{C_n\}$. Compared with directly applying the KMeans clustering~\cite{ahmed2020k} to $\{\mathbf{e}_k\}$, this approach does not require predefining the number of clusters.

\par
We design a hierarchical key events selection mechanism as follows. For each cluster $C_n \in \mathcal{C}$, we compute its similarity score $r_n$ with the query $q$ by
\[
    r_n = \frac{1}{|C_n|} \sum_{e_k \in C_n} \mathrm{sim}_\text{BGE}(e_k, q).
\]
Then we sort the clusters in $\mathcal{C}$ according to $r_n$ in the descending order, and partition the sorted $\mathcal{C}$ into three subsets, each containing around $|\mathcal{C}|/3$ clusters. Let $\mathcal{C}_\text{high}$, $\mathcal{C}_\text{mid}$ and $\mathcal{C}_\text{low}$ denote the three cluster subsets, and their corresponding event set $\mathcal{E}_\text{high}$, $\mathcal{E}_\text{mid}$ and $\mathcal{E}_\text{low}$, respectively.

\par
Notice that clusters in $\mathcal{C}_\text{low}$ are regraded as with low similarity scores with the query. So we merge all the clusters in $\mathcal{E}_\text{low}$ into a single cluster $C_\text{low}$.

\par
For $\mathcal{C}_\text{high}$ and $\mathcal{C}_\text{mid}$, we use the HDBSCAN clustering~\cite{mcinnes2017hdbscan} on $\mathcal{E}_\text{high}$ and $\mathcal{E}_\text{mid}$, respectively, to obtain another round clustering results. Let $R$ denote the HDBSCAN clustering radius parameter. We set a smaller $R=2$ for $\mathcal{E}_\text{high}$ in order to produce more clusters, and a larger $R=4$ for $\mathcal{E}_\text{mid}$. This is to respect that clusters in $\mathcal{C}_\text{high}$ are with more semantically similar events to the query.

\par
After the HDBSCAN clustering on respective $\mathcal{E}_\text{high}$ and $\mathcal{E}_\text{mid}$, together with $C_{\text{low}}$, let $\mathcal{G}=\{G_m\}$ denote the set of newly formed clusters and each $G_m$ is associated with an event set $\mathcal{E}_m$. For each event $e_j \in \mathcal{E}_m$, we propose \textit{intra-cluster centrality} as follows:
\[
    c_j = \frac{1}{|\mathcal{E}_m|-1} \sum_{e_i \in \mathcal{E}_m, e_i \not= e_j}\mathrm{cosine}(\mathbf{e}_i, \mathbf{e}_j).
\]
$c_j$ measures how semantically close event $e_j$ is, on average, to all other events in the same cluster.$e_j$ with the maximum $c_j$ lies at the semantic core of the $G_m$ and serves as its most representative event.
For each cluster $G_m$, we select the event  with the largest intra-cluster centrality as its key event. 

\par
\paragraph{Coreference Resolution: } Let $\mathcal{E}_\text{key}$ denote the set of extracted key events in the previous step. We further perform a coreference resolution on events in $\mathcal{E}_\text{key}$ to obtain the final event summary $S$.  Let $\mathrm{sim}_{\text{GTE}}(\cdot,\cdot)$ denote the GTE-based semantic similarity~\cite{li2023towards} and set a coreference threshold $\theta=0.90$, such that two events are considered coreferential if their similarity exceeds this threshold. Following the order of events in $\mathcal{E}_{\text{key}}$, an event
$e_j \in \mathcal{E}_{\text{key}}$ is added to the final summary $S$
only if it is not semantically redundant with previously selected ones,
i.e.,
\[
\mathrm{sim}_{\text{GTE}}(e_j, e') < \theta, \quad \forall e' \in S.
\]
The final event summary $S$ is obtained after all key events have been examined.
We perform coreference resolution on key events at the final stage, rather than for all extracted events. The rationale is that repeated mentions of semantically similar extracted events also serve as an important signal for event salience, which is a key criterion during event clustering and key events selection.

\section{Experiments}
\subsection{Experimental Settings}

\textbf{Dataset}: All experiments are conducted on QFESum dataset. We release QFESum, together with the implementation of our framework, at \url{https://github.com/sarcasm-hcy02/QFES-QFESum}.

\textbf{Baselines}: Since our proposed method follows a ``retrieve-then-summarize'' paradigm, we separately compare RAT and SHC with strong retrieval and summarization baselines. \textbf{Retrieval}: BM25~\cite{robertson2009probabilistic} , DPR~\cite{karpukhin2020dense}, RoBERTa-large-mnli~\cite{liu2019roberta}. More details and settings are shown in Appendix~\ref{sec:appendix retrieval sttings}.
\textbf{Summarization}: Topic\_TLS(2024, ACL-long)~\cite{hu2024moments}, 
GraphRAG(2024, ArXiv)~\cite{edge2024local}, 
FG-RAG(2025, CIKM-short)~\cite{hong2025fg}, 
UnstructBase(2025, EMNLP-main)~\cite{wright-etal-2025-unstructured}.   
More details and settings are shown in Appendix~\ref{sec:appendix Summarization sttings}

\textbf{Evaluation Metrics}: We employ separate evaluation metrics for the retrieval and summarization components.
\textbf{Retrieval}: Recall, Precision, F1, Hit@5, and Precision@5.
\textbf{Summarization}: Lexical overlap metrics including ROUGE-1/2/L/4~\cite{lin2004rouge}, BLEU~\cite{papineni2002bleu}, and METEOR~\cite{banerjee2005meteor}, as well as a semantic similarity metric BERTScore~\cite{zhang2019bertscore} and LLM-based evaluation metric: LLM-Score.
Appendix~\ref{sec:Detailed Experiment Settings} provides more details on the full experimental settings.

\subsection{Summarization of Gold Documents $\mathcal{D}_{i}^\text{r}$}
\label{sec: std sum result}
For each query $q_i$, we define $q_i$ annotated relevant document set as gold documents $\mathcal{D}_{i}^\text{r}$. In this section, the input consists of the gold document set $\mathcal{D}_{i}^\text{r}$. We compare SHC with baseline summarizers, and the results are provided in Table~\ref{tab:std sum results}. We present the event number of different summarization in Table~\ref{tab:summary_event_num}.

\begin{table*}[t]
\centering
\resizebox{\textwidth}{!}{
\begin{tabular}{lccccccc}
\hline
\textbf{Methods + Input Documents} &
\textbf{ROUGE-1} &
\textbf{ROUGE-2} &
\textbf{ROUGE-L} &
\textbf{ROUGE-4} &
\textbf{BLEU} &
\textbf{METEOR} &
\textbf{BERTScore} \\
\hline
\textbf{SHC}(OURS) + $\mathcal{D}^\text{r}_i$ & \textbf{38.71} & \textbf{10.44} & \textbf{16.08} & \textbf{1.47} & \textbf{4.96} & \textbf{26.75} & \textbf{83.47} \\
Topic\_TLS(2024, ACL) + $\mathcal{D}^\text{r}_i$ & 29.98 & 6.35 & 12.53 & 0.41 & 2.38 & 25.56 & 82.21 \\
GraphRAG + $\mathcal{D}^\text{r}_i$ & 27.76 & 4.85 & 13.72 & 0.18 & 1.67 & 20.32 & 81.34 \\
FG-RAG(2025, CIKM) + $\mathcal{D}^\text{r}_i$ & 27.13 & 6.58 & 12.61 & 0.50 & 2.15 & 15.06 & 82.16 \\
UnstructBase(2025, EMNLP) + $\mathcal{D}^\text{r}_i$ & 27.54 & 4.84 & 12.51 & 0.73 & 2.31 & 17.62 & 81.01 \\
\hline
\end{tabular}
}
\caption{Results of gold documents $\mathcal{D}^\text{r}_i$ summarization}
\label{tab:std sum results}
\end{table*}

It can be observed that SHC significantly outperforms the other QFS methods. Topic\_TLS yields inferior performance, because its LLM-based similarity linking tends to form overly large clusters of semantically related but redundant sentences, resulting in lengthy summaries with lower information density.

The performance gap where GraphRAG and FG-RAG underperform our method can be attributed to they construct graphs directly from source documents, leading to the incorporation of substantial query-irrelevant information during graph construction. In contrast, SHC first extracts structured events from the source documents, thereby reducing the amount of query-irrelevant information. This issue is exacerbated at scale, as longer inputs lead to clear performance degradation. For the UnstructBase Method, its poor performance is mainly due to the excessive amount of extracted evidence in the large-scale dataset. When the LLM is used for summarization, this often leads to severe instruction-following errors and content-selection failures, causing the generated summaries to retain substantial background information and subjective comments that do not meet the requirements of event-oriented summarization. More results for each thematic corpus in the QFESum dataset are shown in Appendix~\ref{sec:Detailed_std_sum_Results}.

\subsection{LLM-based Evaluation Results}
\label{sec:LLM-score}
LLM-score:
In event summarization, a sentence in summary represents an summarized event. The key to the quality of a summary lies in whether a generated event can accurately match an event semantics in reference summary. Accordingly, we introduce an LLM-based event alignment evaluation for QFES,
which computes Precision (LLM-Pre), Recall (LLM-Rec), and F1 (LLM-F1).The prompt template follows~\cite{sun2024towards}, with detailed prompt, computational process shown in Appendix~\ref{sec:LLM Evaluation Settings}. For our experiments, we use two large language models: LLM1 (DeepSeek-v3~\cite{liu2024deepseek}) and LLM2 (Qwen3-32B~\cite{yang2025qwen3}).

\begin{table*}[t]
\centering
\small
\resizebox{\textwidth}{!}{
\begin{tabular}{lcccccc}
\hline
\textbf{Methods + Input Documents} &
\textbf{LLM1-Pre} &
\textbf{LLM1-Rec} &
\textbf{LLM1-F1} &
\textbf{LLM2-Pre} &
\textbf{LLM2-Rec} &
\textbf{LLM2-F1} \\
\hline
\textbf{SHC}(OURS) + $\mathcal{D}^\text{r}_i$ & \textbf{11.63} & 15.27 & \textbf{12.21} & \textbf{21.10} & 28.69 & \textbf{22.63} \\
Topic\_TLS + $\mathcal{D}^\text{r}_i$ & 2.88 & 6.82 & 3.56 & 9.00 & 17.30 & 9.70 \\
GraphRAG + $\mathcal{D}^\text{r}_i$ & 7.93 & 10.73 & 8.63 & 13.26 & 17.99 & 13.17 \\
FG-RAG + $\mathcal{D}^\text{r}_i$ & 7.87 & \textbf{21.56} & 10.71 & 16.67 & \textbf{43.00} & 22.47 \\
UnstructBase + $\mathcal{D}^\text{r}_i$ & 2.48 & 2.67 & 1.92 & 7.12 & 7.09 & 5.56 \\
\hline
\end{tabular}
}
\caption{
LLM-based evaluation containing Precision, Recall, and F1 across baselines.
}
\label{tab:llm_metrics}
\end{table*}

 As provided in Table~\ref{tab:llm_metrics}, compared with Topic\_TLS, our method achieves more stable and precise event summarization. Topic\_TLS often produces too many clusters, which introduces redundancy and leads to lower precision and recall. When comparing with FG-RAG, which achieves higher recall primarily because it generates substantially longer summaries that cover a larger number of events.
However, the inclusion of many irrelevant events leads to lower precision, resulting in the LLM-F1 score that remains inferior to ours.
Moreover, excessively long summaries increase redundancy and reduce conciseness, which is undesirable in query-focused event summarization.

Additionally, LLM evaluators tend to interpret broad or high-level generated events as semantically consistent with reference events, even when such generated events lack specific event details. FG-RAG often produces such generic descriptions, and its generated events are therefore more likely to be judged as matching the reference events, resulting in higher LLM recall. Whereas our summaries focus more on concrete, fact-specific events. More results are shown in Appendix~\ref{sec:Detailed_std_sum_Results}. A case study is shown in Appendix~\ref{sec:CaseStudy} to illustrate a LLM-based evaluation example.



\subsection{Human Evaluation}

\begin{table}[t]
\centering
\scriptsize
\setlength{\tabcolsep}{5pt}
\renewcommand{\arraystretch}{0.95}
\begin{tabular}{lccc}
\hline
\textbf{Method + Input Documents} &
\textbf{Event-P} &
\textbf{Event-R} &
\textbf{Event-F1} \\
\hline
\textbf{SHC} + $\mathcal{D}^{\mathrm{r}}_i$ 
& \textbf{26.88} & \textbf{42.02} & \textbf{29.50} \\
FG-RAG + $\mathcal{D}^{\mathrm{r}}_i$ 
& 12.47 & 32.62 & 16.82 \\
\hline
\end{tabular}
\caption{Human evaluation results between our method and FG-RAG.}
\label{tab:human_evaluation}
\end{table}

To validate the effectiveness of our summarization method and the LLM-based evaluation, we invite three postgraduate annotators  to conduct a human evaluation on the summaries of eight queries under the \texttt{Yemen} topic. Specifically, we compare the summaries generated by our method with those produced by FG-RAG. Following the same calculation procedure as the LLM-based evaluation, we compare each generated summary sentence against the reference summary to determine whether it can be matched to a corresponding event, based on which Event-Pre, Event-Rec, and Event-F1 are calculated. The averaged results across the three annotators are presented in Table~\ref{tab:human_evaluation}.

The human evaluation shows that more than half of the content in the summaries generated by FG-RAG is irrelevant, with a large proportion consisting of background information, introductory descriptions, or other expressions that are not directly related to the query-specific events. Although the summaries generated by FG-RAG are generally longer, their recall under human evaluation is still lower than that of our SHC method, and the gap in precision is even more substantial. These results are consistent with the trend observed in the LLM-based evaluation, suggesting that LLMs can effectively approximate human judgment in event summary evaluation. Meanwhile, the comparison also demonstrates the superiority of our SHC method.

\subsection{Summarization of Retrieved Documents}
\label{sec:retrieved results}

Retrieval results of different retrieval are provided in Table~\ref{tab:doc_retrieve_results}. Summarization results of documents from different retrieval are presented in Table~\ref{tab:retrieved doc sum results}. It should be noted that here, and unless otherwise specified, all reported LLM-F1 scores refer to results evaluated by DeepSeek-v3 model.

\begin{table*}[h]
\centering
\resizebox{\textwidth}{!}{
\begin{tabular}{lcccccccc}
\hline
\textbf{Methods + Input Documents} &
\textbf{ROUGE-1} &
\textbf{ROUGE-2} &
\textbf{ROUGE-L} &
\textbf{ROUGE-4} &
\textbf{BLEU} &
\textbf{METEOR} &
\textbf{BERTScore} &
\textbf{LLM-F1}  \\
\hline

\rowcolor{gray!15}\multicolumn{9}{c}{\textbf{OURS + $\mathcal{D}^\text{r}_i$}} \\
\hline
SHC + \textbf{$\mathcal{D}^\text{r}_i$} & \textbf{38.71} & \textbf{10.44} & \textbf{16.08} & \textbf{1.47} & \textbf{4.96} & \textbf{26.75} & \textbf{83.47} & \textbf{12.21}\\
\hline

\rowcolor{gray!15}\multicolumn{9}{c}{\textbf{OURS + Retrieved Documents}} \\
\hline
SHC + \textbf{RAT}(OURS) & \textbf{35.14} & \textbf{8.76} & \textbf{14.72} & \textbf{0.96} & 3.64 & \textbf{24.87} & 83.25 & \textbf{8.32}\\
SHC + BM25 & 34.55 & 7.90 & 14.70 & 0.77 & \textbf{3.94} & 22.58 & \textbf{83.43} & 5.83\\
SHC + DPR & 32.02 & 7.64 & 13.68 & 0.93 & 3.66 & 22.30 & 83.03 & 5.80\\
SHC + Roberta-mnli & 22.75 & 4.90 & 10.88 & 0.75 & 1.92 & 12.56 & 82.90 & 3.37\\
\hline

\end{tabular}
}
\caption{Summarization results of retrieved documents}
\label{tab:retrieved doc sum results}
\end{table*}

In the retrieval results, RAT achieves the best performance. Benefiting from its automatically learned dynamic threshold rather than manually set fixed hyperparameters, RAT adapts to different semantic distributions across corpus, maintaining a more stable balance between recall, precision and F1, demonstrating strong robustness and adaptability in document retrieval.

It is noteworthy that, under certain queries in our dataset, the simplest method, BM25 outperforms retrieval than DPR and RoBERTa-large-mnli. The underlying reason is that some queries (e.g., \texttt{``impact on wildlife and ecosystems''}) contain highly observable semantic anchors \texttt{"wildlife"} and \texttt{"ecosystems"}, and documents contain such semantic anchors are likely to be labeled as relevant. Therefore, BM25 enjoys a natural advantage in these lexical-match–friendly settings.

\begin{table}[t]
\centering
\small
\setlength{\tabcolsep}{4.8pt}
\renewcommand{\arraystretch}{1.02}
\begin{tabular}{lccccc}
\hline
\textbf{Retrieval Method} &
\textbf{P} &
\textbf{R} &
\textbf{F1} &
\textbf{Hit@5} &
\textbf{P@5} \\
\hline
\textbf{RAT} 
& \textbf{0.6238} & \textbf{0.6710} & \textbf{0.5628} & \textbf{0.9844} & \textbf{0.8065} \\
BM25 
& 0.6215 & 0.4323 & 0.4011 & \textbf{0.9844} & 0.7781 \\
DPR 
& 0.5389 & 0.4526 & 0.3796 & 0.8335 & 0.5259 \\
RoBERTa-MNLI 
& 0.5860 & 0.0466 & 0.0837 & 0.7993 & 0.4858 \\
\hline
\end{tabular}
\caption{Retrieval results of different methods.}
\label{tab:doc_retrieve_results}
\end{table}

In the summarization experiments, overall summary quality exhibits a positive correlation with retrieval performance.
Methods with stronger retrieval effectiveness, RAT followed by BM25, DPR, and RoBERTa-large-mnli—consistently yield better generation results, reinforcing the critical role of retrieval in query-focused event summarization. Meanwhile, BM25 slightly surpasses RAT on BLEU and BERTScore, mainly because its generated summaries have lengths closer to the reference and exhibit stronger semantic concentration in the input. More results are provided in Appendix~\ref{sec:Detailed retrieved sum Results}.

\subsection{Ablation Studies}
\label{tac:section ablation}
We conduct ablation studies on two key modules: the hierarchical clustering mechanism in \texttt{Key Events Selection}  and the retrieval module. Specifically, removing the hierarchical mechanism means that in \texttt{Key Events Selection} after BERTopic clustering, we perform HDBSCAN clustering without semantic hierarchies; Removing the retrieval module means inputting extracted events from all documents. The ablation results are presented in Table~\ref{tab:ablation_results}.

It can be observed that the hierarchical mechanism plays a crucial role in SHC: removing hierarchical clustering mechanism leads to notable degradation on all metrics except BERTScore. In addition, comparing ``w/o retrieval'' versus ``w/o retrieval + w/o hierarchies'' further confirms the effectiveness of the hierarchical design: By reducing intra-cluster redundancy and improving the balance of thematic distribution, our hierarchical clustering mechanism significantly enhances both focus and coverage.

\begin{table}[t]
\centering
\scriptsize
\setlength{\tabcolsep}{4.8pt}
\renewcommand{\arraystretch}{1.02}
\begin{tabular}{lccccc}
\hline
\textbf{Method} &
\textbf{R-1} &
\textbf{R-2} &
\textbf{R-L} &
\textbf{BS} &
\textbf{LLM-F1} \\
\hline
\textbf{SHC} + $\mathcal{D}^{\mathrm{r}}_i$  
& \textbf{38.71} & \textbf{10.44} & \textbf{16.08} & 83.47 & \textbf{12.21} \\
SHC + RAT 
& 35.14 & 8.76 & 14.72 & 83.25 & 8.32 \\
w/o hier. + $\mathcal{D}^{\mathrm{r}}_i$ 
& 34.93 & 8.61 & 14.85 & \textbf{83.67} & 7.26 \\
SHC w/o retrieval 
& 33.09 & 8.63 & 13.66 & 83.26 & 6.98 \\
w/o retrieval \& hier. 
& 31.75 & 8.04 & 13.26 & 83.15 & 5.86 \\
\hline
\end{tabular}
\caption{Ablation study results.}
\label{tab:ablation_results}
\end{table}

In the ``w/o retrieval'' setting, where the model extracts events from all documents rather than from gold ones, summarization quality drops markedly on all metrics. This aligns with the role of retrieval in noise reduction. Interestingly, the performance gap between ``w/o retrieval'' and RAT retrieval is smaller than expected. This is because SHC implicitly conducts a semantic filtering during \texttt{Event Extraction}, making it partially robust to noisy input; However, without retrieval constraints, the extraction process becomes substantially more costly and error-prone, highlighting the continued importance of high-quality retrieval. More detailed results are provided in Appendix~\ref{sec:Detailed ablation Results}.

\section{Conclusion and Future Work}
In this work, we have introduced the Query-Focused Event Summarization (QFES) task and constructed QFESum , a large-scale QFES dataset. Moreover, we propose a two-component summarization framework consisting of Query-Focused Retrieval with Adaptive Thresholding RAT and Query-Focused Summarizer based on Hierarchical Clustering SHC. Extensive experiments demonstrate the effectiveness of our approach in complex thematic corpora, providing new directions for future research in QFES. In the next step, we will explore algorithms that integrate knowledge graphs with QFES.

\section*{Acknowledgments}

The authors thank the anonymous reviewers and the Area Chair for their valuable feedback and suggestions. 

\section*{Limitations}

This study is subject to notable computational constraints throughout experimentation, the majority of model inference and automatic annotation procedures were conducted using Qwen2.5-7B. Its capacity remains inferior to larger-scale models, which may impose performance limitations. Nevertheless, since the annotated labels are derived based on reference summaries, their reliability in capturing true query relevance is substantially higher than retrieval relying solely on the query text. Therefore, employing these annotations as gold documents for evaluating retrieval quality remains feasible and justified.

\section*{Ethics Statement}

This study exclusively utilizes publicly accessible
datasets, and all models employed are publicly accessible . 

\bibliographystyle{plain}
\bibliography{QFES_arXiv}

\appendix

\section{Related Work}

In this section, we first review related work on QFS datasets, followed by existing studies on QFS methods.

\subsection{QFS Datasets}
Early widely used datasets include DUC~\cite{dang2005overview} and TAC~\cite{dang2008overview}, both released by NIST to support summarization research. These datasets contain multiple news articles with queries spanning diverse domains such as politics, biography, and disaster events. Although they are targeted at query-focused summarization within thematic corpora, access to these datasets currently requires formal application, making them less convenient for research. Moreover, their limited number of documents and queries prevents them from forming large-scale resources; summaries constructed from only a few dozen documents struggle to comprehensively capture the narrative of thematic events. A dataset with similar scale constraints is TD-QFS~\cite{baumel2016topic}, a medical dataset covering four topics (asthma, lung cancer, obesity, and Alzheimer's disease), where the queries focus on factors such as causes, treatments, and adolescent impact.

In recent years, research interest in QFS datasets has increased, motivated by users’ desire for personalized and query-driven information access. However, few newly proposed QFS datasets are explicitly designed for event summarization. For example, AQuaMuSe~\cite{kulkarni2020aquamuse} simulates search engines generating high-quality paragraphs for user queries based on NQ~\cite{kwiatkowski2019natural} and Common Crawl, resulting in 5,519 query-focused summaries with an average of 6 documents each. QMSum~\cite{zhong2021qmsum} focuses on query-based multi-domain meeting summarization, containing 1,808 summaries from 232 meetings across academic, product, and committee domains. SQuALITY~\cite{wang2022squality}, built on QuALITY~\cite{pang2022quality}, is a question-driven dataset of public-domain short stories, including 100 stories with 5 document–query pairs per story, each 3k–6k words in length. WebCiteS~\cite{deng2024webcites} is a Chinese dataset designed for attributed query-focused summarization (AQFS), containing 7k human-annotated summaries with citations.However, these datasets are not explicitly designed around events, lack document–query relevance labels, or contain too few documents to ensure comprehensive event coverage—thereby limiting their suitability for real-world, QFES research.

\subsection{QFS Methods}
Query-focused summarization (QFS) aims to generate summaries that stay highly relevant to a given query conditioned on a set of input documents. Early work in this area predominantly adopted unsupervised extractive approaches. For example, Wan et al.~\cite{wan2007manifold} introduced a manifold-ranking–based sentence extraction method. With the advent of Transformer architectures~\cite{vaswani2017attention}, Transformer-based summarizers have achieved substantial performance gains~\cite{roy2023review}. In this stage of development, various techniques were further introduced to enhance summary quality. SEGENC~\cite{vig2022exploring} enables the decoder to attend jointly to multiple encoded segments using segment encoders. Socratic~\cite{pagnoni2023socratic} augments SEGENC with question-driven pretraining to strengthen query grounding. Ranking-based learning has also become widely adopted: Ranker-Generator~\cite{liu2023learning} leverages pairwise and global ranking to select the top-$k$ most relevant content before generation, while Sotudeh et al.~\cite{sotudeh2024learning} jointly train a decoder for both ranking and summarization to improve paragraph prioritization. Nath et al.~\cite{nath2023reinforcement} further introduced reinforcement learning (RL) into QFS, training multiple policy gradient agents based on diverse reward signals.

Beyond the above techniques, graph-based approaches have gained growing attention due to their ability to capture rich structural relations across documents. GraphRAG~\cite{edge2024local} first constructs a complete knowledge graph from text, performs community detection, then generates hierarchical summaries in a bottom-up fashion. LightRAG~\cite{guo2024lightrag} combines graph structure with vector representation in a dual-retrieval framework to improve semantic coverage. NaiveRAG~\cite{gao2023retrieval} segments documents into chunks and retrieves the most relevant ones using dense embeddings. SubGraphRAG~\cite{li2024simple} enhances retrieval precision via entity-based subgraph expansion, particularly benefiting multi-hop QA settings. FG-RAG~\cite{hong2025fg} further introduces context-aware entity expansion to increase entity coverage within the graph and provide richer supporting context.However, dense semantic retrieval methods are limited by PLM input windows, while graph-based generative models incorporate large amounts of query-irrelevant content, causing noise and overly generic summaries as corpus size grows—revealing a clear gap in scaling QFS to long-document and real-world thematic event settings.

\section{Case Study}
~\label{sec:CaseStudy}

\begin{figure*}[!t]
    \centering
    \includegraphics[width=\textwidth]{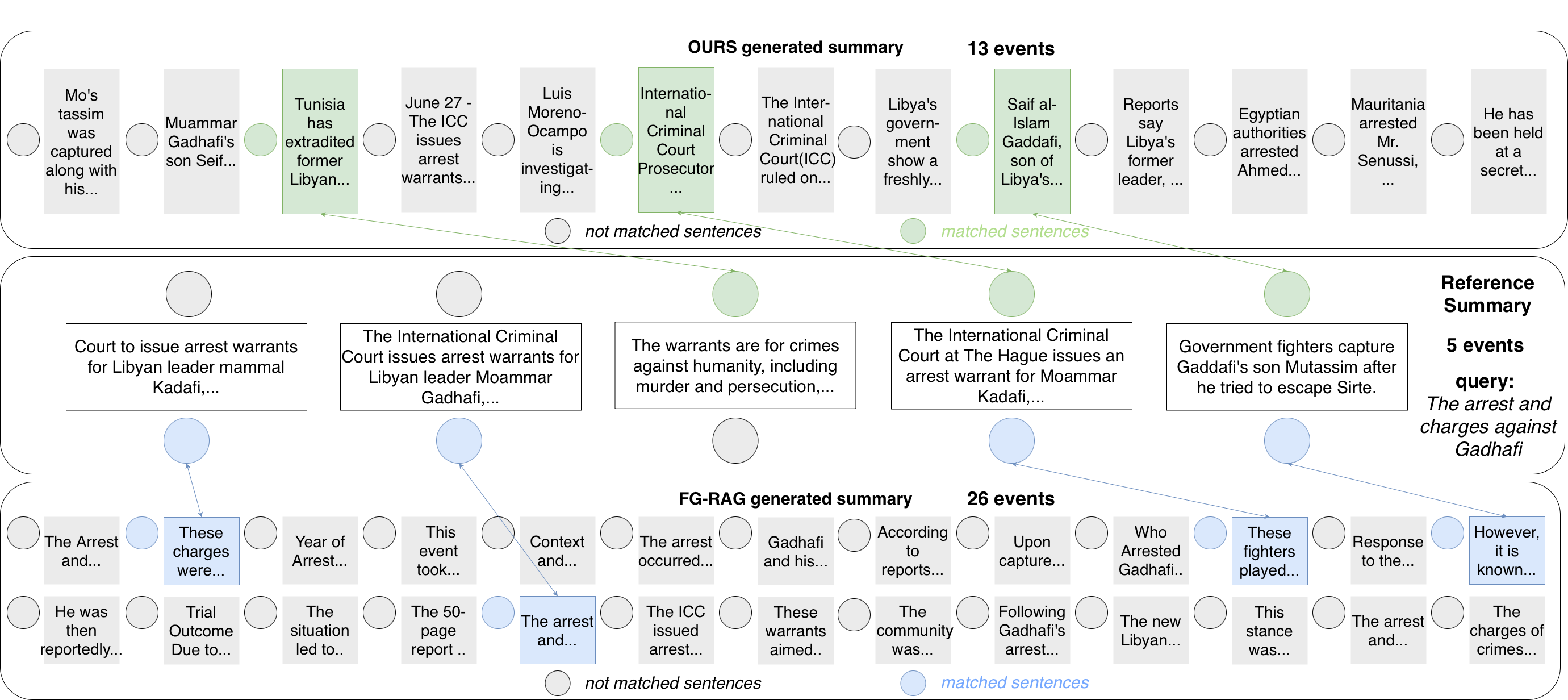}
    \caption{Case study of LLM-based evaluation under the query \texttt{"The arrest and charges against Gadhafi"}.}
    \label{pic:case-study}
\end{figure*}

As shown in Figure~\ref{pic:case-study}, this section presents a case study to illustrate large language model–based evaluation examples of our method and FG-RAG when performing on the QFESum dataset. Under a query \texttt{"The arrest and charges against Gadhafi"} in the Libya thematic corpus, the reference summary contains a total of 5 sentences. Each sentence corresponds to one event, resulting in a total of five events. Our method generates 13 events. The LLM iterates over the 13 generated events and, for each event, compares it against the 5 events in the reference summary to determine semantic matches, ensuring that each reference event is matched at most once. Among our generated events, 3 events are judged by the LLM to have semantic correspondences with the reference summary, as highlighted with \textcolor{green}{green circles}. Accordingly, LLM-Pre = 23.08\%, LLM-Rec = 60.00\%, and LLM-F1 = 33.34\%.

In contrast, the FGRAG generates a longer summary, with 26 events in total, of which 4 events are matched as highlighted with \textcolor{blue}{blue circles}, reflecting a stronger recall but lower precision, LLM-Pre = 15.38\%, LLM-Rec = 80.00\%, and LLM-F1 = 25.80\% .

\section{Detailed Experiment Results}
\label{sec:Detailed Experiment Results}

QFESum consists of eight thematic events: \texttt{\textbf{BPoil}} (BP Oil Spill, from T17), \texttt{\textbf{finan}} (Global Financial Crisis, from T17), \texttt{\textbf{iraq}} (Iraq War, from T17), \texttt{\textbf{syria\_t17}} (Syrian Conflict, from T17), \texttt{\textbf{egypt}} (Egyptian Revolution, from CRISIS), \texttt{\textbf{libya}} (Libyan Civil War, from CRISIS), \texttt{\textbf{syria\_crisis}} (Syrian Conflict, from CRISIS), and \texttt{\textbf{yemen}} (Yemeni Uprising, from CRISIS). This section provides detailed per-corpus evaluation results for all experiments.

\subsection{Detailed Gold Documents Summarization Results}
\label{sec:Detailed_std_sum_Results}

Table~\ref{tab:summary_event_num} reports the  number of events of different summarization methods.
Table~\ref{tab:detailed_standard_results} reports ROUGE-1/2/L, BERTScore, and LLM-F1 for the gold document summarization. Here, LLM1-F1 denotes scores evaluated by DeepSeek-v3, while LLM2-F1 denotes scores evaluated by Qwen3-32B.

\begin{table}[t]
\centering
\scriptsize
\setlength{\tabcolsep}{5pt}
\renewcommand{\arraystretch}{0.95}
\begin{tabular}{lc}
\hline
\textbf{Method + Input Documents} & \textbf{Avg. \#Evt.} \\
\hline
Reference Summary & 21.53 \\
SHC + $\mathcal{D}^{\mathrm{r}}_i$ & 28.01 \\
TopicTLS + $\mathcal{D}^{\mathrm{r}}_i$ & 40.48 \\
GraphRAG + $\mathcal{D}^{\mathrm{r}}_i$ & 16.28 \\
FG-RAG + $\mathcal{D}^{\mathrm{r}}_i$ & 47.01 \\
UnstructBase + $\mathcal{D}^{\mathrm{r}}_i$ & 17.69 \\
\hline
\end{tabular}
\caption{Average number of summary events generated by different methods.}
\label{tab:summary_event_num}
\end{table}

\begin{table*}[t]
\centering
\small
\resizebox{\textwidth}{!}{
\begin{tabular}{llcccccc}
\toprule
\textbf{Methods + Input Documents} &
\textbf{Corpus} &
\textbf{ROUGE-1} &
\textbf{ROUGE-2} &
\textbf{ROUGE-L} &
\textbf{BERTScore} &
\textbf{LLM1-F1} &
\textbf{LLM2-F1} \\
\midrule
\multirow{8}{*}{\textbf{SHC} + $\mathcal{D}^\text{r}_i$}
 & BPoil & 38.21 & 10.68 & 16.1 & 83.53 & 10.15 & 23.76 \\
 & finan & 32.95 & 7.45 & 13.64 & 82.23 & 10.21 & 21.09 \\
 & iraq & 39.37 & 10.28 & 17.88 & 82.66 & 11.45 & 19.34 \\
 & syria\_t17 & 40.02 & 11.73 & 17.09 & 84.31 & 16.85 & 23.04 \\
 & egypt & 40.07 & 8.70 & 15.42 & 83.26 & 16.91 & 22.71 \\
 & libya & 38.04 & 10.37 & 16.19 & 83.99 & 7.09 & 17.09 \\
 & syria\_crisis & 42.39 & 14.16 & 16.42 & 84.29 & 13.20 & 23.59 \\
 & yemen & 42.47 & 12.24 & 17.74 & 84.24 & 11.82 & 31.42 \\
\midrule
\multirow{8}{*}{Topic\_TLS + $\mathcal{D}^\text{r}_i$}
 & BPoil & 32.14 & 8.51 & 14.03 & 82.59 & 6.95 & 13.46 \\
 & finan & 33.40 & 6.66 & 13.40 & 82.13 & 3.27 & 9.41 \\
 & iraq & 29.69 & 5.75 & 12.29 & 81.59 & 1.14 & 6.28 \\
 & syria\_t17 & 28.02 & 5.08 & 11.61 & 81.96 & 2.56 & 10.46 \\
 & egypt & 26.06 & 5.02 & 11.56 & 81.94 & 2.42 & 8.81 \\
 & libya & 26.69 & 6.37 & 11.12 & 82.40 & 3.02 & 8.60 \\
 & syria\_crisis & 32.53 & 6.82 & 13.01 & 82.51 & 5.01 & 9.26 \\
 & yemen & 31.29 & 6.60 & 13.24 & 82.58 & 4.14 & 11.32 \\
\midrule
\multirow{8}{*}{GraphRAG + $\mathcal{D}^\text{r}_i$}
 & BPoil & 29.93 & 6.08 & 15.13 & 80.41 & 5.26 & 17.47 \\
 & finan & 28.10 & 4.82 & 13.64 & 80.44 & 7.11 & 14.01 \\
 & iraq & 30.25 & 5.59 & 14.59 & 81.61 & 12.11 & 14.01 \\
 & syria\_t17 & 28.72 & 5.05 & 13.83 & 82.26 & 20.58 & 12.78 \\
 & egypt & 24.54 & 3.14 & 12.37 & 80.75 & 5.02 & 7.61 \\
 & libya & 25.51 & 4.20 & 13.43 & 81.06 & 3.58 & 6.71 \\
 & syria\_crisis & 30.60 & 5.54 & 14.29 & 82.77 & 10.72 & 19.91 \\
 & yemen & 24.39 & 4.35 & 12.51 & 81.40 & 4.68 & 12.89 \\
\midrule
\multirow{8}{*}{FG-RAG + $\mathcal{D}^\text{r}_i$}
 & BPoil & 31.16 & 8.75 & 14.43 & 82.05 & 10.72 & 23.63 \\
 & finan & 27.31 & 5.80 & 12.72 & 80.84 & 8.05 & 16.70 \\
 & iraq & 31.32 & 8.15 & 14.32 & 82.40 & 13.76 & 27.20 \\
 & syria\_t17 & 25.89 & 6.30 & 11.92 & 82.50 & 11.55 & 20.45 \\
 & egypt & 24.39 & 5.76 & 11.63 & 81.96 & 11.99 & 21.96 \\
 & libya & 24.60 & 5.84 & 11.80 & 82.49 & 8.43 & 18.69 \\
 & syria\_crisis & 28.64 & 6.53 & 13.10 & 83.05 & 11.48 & 27.76 \\
 & yemen & 23.70 & 5.54 & 10.97 & 81.99 & 9.66 & 23.33 \\
 \midrule
\multirow{8}{*}{UnstructBase + $\mathcal{D}^\text{r}_i$}
 & BPoil & 27.16 & 6.11 & 13.32 & 80.72 & 3.68 & 9.90 \\
 & finan & 22.07 & 2.99 & 10.26 & 79.34 & 0.50 & 3.50 \\
 & iraq & 26.37 & 7.72 & 14.75 & 80.65 & 4.16 & 7.99 \\
 & syria\_t17 & 25.85 & 4.71 & 11.11 & 81.86 & 0.00 & 2.29 \\
 & egypt & 38.26 & 5.27 & 14.29 & 82.05 & 0.77 & 3.49 \\
 & libya & 23.10 & 3.67 & 10.70 & 80.33 & 3.41 & 6.47 \\
 & syria\_crisis & 28.78 & 4.40 & 13.34 & 82.08 & 0.66 & 4.67 \\
 & yemen & 28.69 & 3.81 & 12.32 & 81.03 & 2.16 & 6.17 \\
\bottomrule
\end{tabular}
}
\caption{
Per-corpus evaluation of five Methods + Gold Documents on 8 corpus,
evaluated by ROUGE-1, ROUGE-2, ROUGE-L, BERTScore, LLM1-F1 and LLM2-F1.
}
\label{tab:detailed_standard_results}
\end{table*}

\subsection{Detailed Retrieved Documents Summarization Results}
\label{sec:Detailed retrieved sum Results}

Table~\ref{tab:detailed_retrieved_results_table} presents summarization and retrieval results from different retrieved methods, corresponding to the experiments described in Section~\ref{sec:retrieved results}. Due to space limitations, we report the following evaluation metrics: ROUGE-1/2/L and LLM-F1 (where LLM-F1 is computed using DeepSeek-v3) for summarization performance, and F1 and Hit@5 for retrieval performance.

\begin{table*}[t]
\centering
\small
\resizebox{\textwidth}{!}{
\begin{tabular}{llcccc|cc}
\toprule
\textbf{Methods + Input Documents} &
\textbf{Corpus} &
\multicolumn{4}{c}{\textbf{Summary Metrics}} &
\multicolumn{2}{c}{\textbf{Retrieval Metrics}} \\
\cmidrule(lr){3-6} \cmidrule(lr){7-8}
 & &
\textbf{ROUGE-1} &
\textbf{ROUGE-2} &
\textbf{ROUGE-L} &
\textbf{LLM-F1} &
\textbf{F1} &
\textbf{hit@5} \\
\midrule
\multirow{8}{*}{SHC + \textbf{RAT}}
 & BPoil         & 40.67 & 11.81 & 17.10 & 12.60 & 0.4952 & 1.0000 \\
  & finan         & 27.56 & 5.73  & 11.82 & 2.29  & 0.5486 & 1.0000 \\
  & iraq          & 32.08 & 6.09  & 13.77 & 4.04  & 0.5564 & 1.0000 \\
  & syria\_t17    & 34.35 & 8.62  & 14.85 & 10.57 & 0.8152 & 1.0000 \\
  & egypt         & 39.19 & 8.38  & 14.85 & 6.47  & 0.5801 & 1.0000 \\
  & libya         & 34.44 & 8.88  & 15.27 & 8.36  & 0.4606 & 1.0000 \\
  & syria\_crisis & 35.23 & 10.12 & 14.43 & 10.92 & 0.5134 & 1.0000 \\
  & yemen         & 36.93 & 9.59  & 15.89 & 11.11 & 0.5325 & 0.8750 \\
\midrule
\multirow{8}{*}{SHC + BM25}
 & BPoil         & 38.63 & 10.76 & 16.54 & 9.99  & 0.4789 & 1.0000 \\
  & finan         & 30.49 & 6.95  & 11.94 & 3.47  & 0.5229 & 1.0000 \\
  & iraq          & 33.19 & 6.14  & 14.82 & 2.19  & 0.5089 & 1.0000 \\
  & syria\_t17    & 27.45 & 6.26  & 13.14 & 5.32  & 0.6826 & 1.0000 \\
  & egypt         & 37.20 & 7.64  & 15.05 & 8.95  & 0.2975 & 1.0000 \\
  & libya         & 35.28 & 8.09  & 15.21 & 5.33  & 0.2919 & 1.0000 \\
  & syria\_crisis & 38.00 & 8.75  & 15.40 & 6.63  & 0.2126 & 1.0000 \\
  & yemen         & 36.17 & 8.57  & 15.46 & 4.79  & 0.2133 & 0.8750 \\
\midrule
\multirow{8}{*}{SHC + DPR}
 & BPoil         & 41.27 & 12.07 & 17.07 & 9.42  & 0.3871 & 0.7619 \\
  & finan         & 23.51 & 4.72  & 10.12 & 2.27  & 0.4343 & 0.8095 \\
  & iraq          & 29.96 & 5.45  & 12.99 & 5.15  & 0.3669 & 1.0000 \\
  & syria\_t17    & 18.93 & 3.18  & 10.04 & 1.54  & 0.3179 & 1.0000 \\
  & egypt         & 34.05 & 6.79  & 14.77 & 6.38  & 0.4080 & 0.9000 \\
  & libya         & 35.85 & 8.35  & 14.84 & 4.53  & 0.3334 & 0.5714 \\
  & syria\_crisis & 37.30 & 11.18 & 14.71 & 8.38  & 0.3891 & 1.0000 \\
  & yemen         & 35.32 & 9.41  & 14.88 & 8.76  & 0.4004 & 0.6250 \\
\midrule
\multirow{8}{*}{SHC + Roberta-mnli}
 & BPoil         & 13.45 & 2.36  & 8.36  & 2.22  & 0.0693 & 0.9048 \\
  & finan         & 14.10 & 2.74  & 8.25  & 0.56  & 0.0657 & 0.8571 \\
  & iraq          & 19.48 & 3.50  & 10.16 & 0.00  & 0.0893 & 0.9286 \\
  & syria\_t17    & 11.38 & 2.13  & 6.78  & 0.00  & 0.1496 & 1.0000 \\
  & egypt         & 29.20 & 6.51  & 13.13 & 5.46  & 0.0963 & 0.9000 \\
  & libya         & 39.86 & 11.19 & 15.90 & 13.43 & 0.0811 & 0.4286 \\
  & syria\_crisis & 31.73 & 6.33  & 13.87 & 2.42  & 0.0674 & 0.7500 \\
  & yemen         & 23.63 & 4.47  & 10.78 & 2.90  & 0.0507 & 0.6250 \\
\bottomrule
\end{tabular}
}
\caption{
Per-corpus evaluation across retrieval baselines using Summary Metrics
(ROUGE-1/2/L, LLM-F1) and Retrieval Metrics (F1, hit@5).
}
\label{tab:detailed_retrieved_results_table}
\end{table*}

\subsection{Detailed Ablation Studies Results}
\label{sec:Detailed ablation Results}

Table~\ref{tab:Detailed Ablation Results} presents the detailed results of the ablation study. We report metrics including ROUGE-1/2/L, BERTScore, and LLM-F1 computed using DeepSeek-v3.

\begin{table*}[t]
\centering
\small
\resizebox{\textwidth}{!}{
\begin{tabular}{llccccc}
\toprule
\textbf{Methods} &
\textbf{Corpus} &
\textbf{ROUGE-1} &
\textbf{ROUGE-2} &
\textbf{ROUGE-L} &
\textbf{BERTScore} &
\textbf{LLM-F1} \\
\midrule

\multirow{8}{*}{SHC + $\mathcal{D}^\text{r}_i$}
  & BPoil & 38.21 & 10.68 & 16.1 & 83.53 & 10.15  \\
 & finan & 32.95 & 7.45 & 13.64 & 82.23 & 10.21  \\
 & iraq & 39.37 & 10.28 & 17.88 & 82.66 & 11.45  \\
 & syria\_t17 & 40.02 & 11.73 & 17.09 & 84.31 & 16.85  \\
 & egypt & 40.07 & 8.70 & 15.42 & 83.26 & 16.91  \\
 & libya & 38.04 & 10.37 & 16.19 & 83.99 & 7.09  \\
 & syria\_crisis & 42.39 & 14.16 & 16.42 & 84.29 & 13.20  \\
 & yemen & 42.47 & 12.24 & 17.74 & 84.24 & 11.82  \\
\midrule

\multirow{8}{*}{SHC + RAT}
 & BPoil         & 40.67 & 11.81 & 17.10 & 84.00 & 12.60 \\
 & finan         & 27.56 & 5.73  & 11.82 & 81.76 & 2.29  \\
 & iraq          & 32.08 & 6.09  & 13.77 & 82.28 & 4.04  \\
 & syria\_t17    & 34.35 & 8.62  & 14.85 & 83.33 & 10.57 \\
 & egypt         & 39.19 & 8.38  & 14.85 & 83.21 & 6.47  \\
 & libya         & 34.44 & 8.88  & 15.27 & 83.62 & 8.36  \\
 & syria\_crisis & 35.23 & 10.12 & 14.43 & 84.27 & 10.92 \\
 & yemen         & 36.93 & 9.59  & 15.89 & 83.50 & 11.11 \\
\midrule

\multirow{8}{*}{w/o hierarchies}
 & BPoil         & 34.21 & 9.19  & 15.59 & 83.32 & 8.75  \\
 & finan         & 29.13 & 6.54  & 12.01 & 82.43 & 3.48  \\
 & iraq          & 26.10 & 4.80  & 12.59 & 82.13 & 2.30  \\
 & syria\_t17    & 37.20 & 9.42  & 15.94 & 84.41 & 7.61  \\
 & egypt         & 36.23 & 6.81  & 14.47 & 83.53 & 8.22  \\
 & libya         & 39.12 & 9.95  & 16.53 & 84.20 & 6.14  \\
 & syria\_crisis & 36.09 & 10.25 & 14.59 & 84.37 & 8.05  \\
 & yemen         & 41.34 & 11.93 & 17.07 & 84.93 & 13.50 \\
\midrule

\multirow{8}{*}{w/o retrieval}
 & BPoil         & 38.81 & 11.32 & 16.15 & 83.53 & 9.61 \\
 & finan         & 29.50 & 7.03  & 11.56 & 82.26 & 2.54 \\
 & iraq          & 34.07 & 7.04  & 14.63 & 82.46 & 3.74 \\
 & syria\_t17    & 33.09 & 9.08  & 14.22 & 83.85 & 6.41 \\
 & egypt         & 36.59 & 7.98  & 13.90 & 83.33 & 11.10 \\
 & libya         & 29.30 & 7.89  & 12.78 & 83.25 & 6.08 \\
 & syria\_crisis & 29.81 & 9.26  & 12.42 & 83.82 & 10.05 \\
 & yemen         & 34.85 & 9.78  & 13.91 & 83.35 & 6.31 \\
\midrule

\multirow{8}{*}{w/o retrieval\&hierarchies}
 & BPoil         & 36.13 & 10.11 & 15.69 & 83.20 & 9.55 \\
 & finan         & 27.53 & 6.63  & 10.81 & 81.85 & 1.81 \\
 & iraq          & 30.20 & 5.18  & 13.23 & 82.11 & 1.39 \\
 & syria\_t17    & 32.82 & 7.18  & 13.48 & 83.63 & 5.94 \\
 & egypt         & 35.28 & 7.47  & 13.72 & 83.20 & 6.02 \\
 & libya         & 31.94 & 8.62  & 13.78 & 83.49 & 5.14 \\
 & syria\_crisis & 26.58 & 8.73  & 11.16 & 83.76 & 8.11 \\
 & yemen         & 33.55 & 10.42 & 14.19 & 83.96 & 8.94 \\
\bottomrule

\end{tabular}
}
\caption{
Ablation study across eight corpus comparing the full model and ablated variants
(w/o hierarchies, w/o retrieval, w/o retrieval\&hierarchies), evaluated on ROUGE-1/2/L,
BERTScore and LLM-F1. }
\label{tab:Detailed Ablation Results}
\end{table*}

\section{Detailed Experiment Settings}
\label{sec:Detailed Experiment Settings}

This section describes the detailed configurations used in our experiments.

\subsection{Experimental Setup}
DeepSeek-v3 and ChatGPT were accessed through API calls, while all other experiments were conducted on a MacBook Pro equipped with an Apple M3 Pro chip and 18GB memory. During dataset construction, constructing each query required approximately 2 hours of human effort and 4 hours of program runtime. During summarization, generating the summary for each query took approximately 4 hours. The dataset contains 104 queries in total.

\subsection{Retrieval Arguments Settings}
\label{sec:appendix retrieval sttings}

In the retrieval stage, our RAT model does not require explicit hyperparameter settings. For comparison baselines, we configure the retrieval modules as follows:

\textbf{BM25} is implemented under a bag-of-words retrieval framework to identify documents semantically relevant to a query in long event-centric collections. Term frequency, document frequency, and inverse document frequency (IDF) are computed over the entire corpus. We adopt the default hyperparameters $k=1.5$ and $b=0.75$ for BM25 scoring. For each query, Top-$K$ documents are retrieved based on BM25 scores, where $K=200$, with a minimum score cutoff of 0.2.

\textbf{Dense Passage Retrieval (DPR)}~\cite{karpukhin2020dense} is implemented using a dual-encoder setup: DPR-question-encoder for queries and DPR-context-encoder for documents (both pretrained on NQ). Documents are split into sliding windows with a maximum length of 512 tokens and stride of 96 for improved coverage. All chunk embeddings are L2-normalized, and the document relevance score is the maximum inner-product (cosine similarity) over all chunks. A threshold-based filtering is applied, where documents with similarity $\geq 0.55$ are considered relevant.

\textbf{RoBERTa-large-MNLI} is used as an NLI-based semantic retrieval model. Long documents are encoded with the same sliding-window strategy (max length 512, stride 96). For each chunk, we construct three “relevant” templates and one “irrelevant” template by pairing the document with the query in natural language form. Predictions are classified into entailment (ENT), neutral (NEU), or contradiction (CONTRA). We adopt a \textit{semantic\_strict} rule: a document is labeled as relevant if \textit{any} chunk entails \textit{any} relevant template (predicted as ENT).

These configurations ensure a fair and consistent evaluation of retrieval effectiveness across all models.

\subsection{Summarization Arguments Settings}
\label{sec:appendix Summarization sttings}

\textbf{Event Extraction Settings:}
As shown in Figure~\ref{pic:prompt_extraction}, we present the Prompt Used for Event Extraction in our framework.

\tcbset{
  promptbox/.style={
    enhanced,
    colback=white,
    colframe=gray!70,
    colbacktitle=gray!30,
    coltitle=black,
    fonttitle=\bfseries,
    boxrule=0.5pt,
    arc=1pt,
    outer arc=1pt,
    left=6pt,
    right=6pt,
    top=6pt,
    bottom=6pt,
    title=#1
  }
}
\begin{figure*}[t]
\centering
\begin{tcolorbox}[promptbox={Prompt for Event Extraction}, width=\linewidth]
\small
\setlength{\parindent}{0pt}
\setlength{\parskip}{2pt}

\textbf{Role:} Event Extraction Assistant

\textbf{General Description.}
You are an Event Extraction assistant. Your task is to extract all events that are strictly relevant to a given query \texttt{\{query\}} from the text. The input document is a news article or report, and the query represents a specific topic of interest. Do not explain your reasoning, only output the final JSON.

\textbf{Instructions:}
\begin{itemize}\setlength{\itemsep}{1pt}\setlength{\topsep}{1pt}\setlength{\partopsep}{0pt}
  \item Read the provided text.
  \item Extract all \textbf{events} that are directly related to the query.
  \item Each event should be a standalone sentence or phrase that captures what happened, who did it, and any consequence if available.
  \item Only include events that are clearly connected to the query.
  \item Event retrieval mainly focuses on the sentence level. You may refine incomplete or overly long sentences using context.
  \item Output the results in the JSON format specified below.
\end{itemize}

\textbf{Output Format:}
\begin{tcblisting}{
  listing only,
  colback=white,
  colframe=gray!40,
  boxrule=0.3pt,
  left=4pt,right=4pt,top=3pt,bottom=3pt,
  listing options={basicstyle=\ttfamily\footnotesize,breaklines=true,columns=fullflexible}
}
{
  "events": [
    "First relevant event.",
    "Second relevant event.",
    ...
  ]
}
\end{tcblisting}

\textbf{Final Instruction.}
Now, use this format to process the following document:

\textbf{Query:} \texttt{\{query\}}

\textbf{Text:} \texttt{\{text\}}

\textbf{Output:}

\end{tcolorbox}
\caption{Prompt for Event Extraction}
\label{pic:prompt_extraction}
\end{figure*}

\textbf{SHC Arguments Settings:}

Due to substantial variations across corpus in terms of the number of original documents (\textit{gold\_documents\_num}), extracted events (\textit{gold\_documents\_event\_num}), and reference summary event number (\textit{reference\_summary\_event\_num})---as presented in Table~\ref{tab:topics_summary_lengths}---the difficulty of the summarization task can differ significantly with respect to compression ratio and information density. To mitigate the influence of such input-scale imbalance on output length, we introduce three controllable parameters during the BERTopic clustering in SHC: \textit{min\_dist}, \textit{min\_cluster\_size}, and \textit{min\_samples}, whose configurations are detailed in Table~\ref{tab:shc_args_settings}. Specifically, \textit{min\_dist}, \textit{min\_cluster\_size}, and \textit{min\_samples} influence semantic compactness and valid cluster formation in the second-stage clustering. 

It is worth emphasizing that parameter tuning in this study is guided by a practical objective---ensuring that the generated summaries remain within a reasonable length range---rather than corpus-specific hyper-optimization. With this lightweight global adjustment, the average generated summary event number across the eight corpus are 22.79, 41.62, 11.14, 20.62, 26.00, 24.71, 38.12, and 25.00 respectively. Although some variations remain, the outputs are overall maintained at a comparable scale, preventing compression-ratio disparity.

\textbf{Summarization Baselines Settings:}

 \textbf{Topic\_TLS}: Clustering events via LLM-based similarity and selects key representatives. We adopt Qwen2.5-7B to extract query-focused events, following the clustering configuration in~\cite{hu2024moments}: for T17, the top 36 clusters are selected and 3 events sampled from each; for CRISIS, the top 29 clusters are selected and 2 events sampled from each.\\
 \textbf{GraphRAG}: Input original documents rather than event list, building a knowledge graph for hierarchical community summarization. Implemented using \texttt{ChatGPT-4o}-mini with all default parameters.\\
  \textbf{FG-RAG}: Augments GraphRAG with context-aware entity expansion to improve retrieved coverage. Implemented using Qwen2.5-7B with all default parameters.\\
  \textbf{UnstructBase}: While proposing this evidence-extraction-based summarization paradigm, the original paper also introduced the SUnSet dataset to improve summarization quality. However, due to resource limitations, this work only follows the paradigm during reproduction and does not conduct additional training; instead, we use the Qwen2.5-7B model directly. Since our method also involves no training procedure, this reproduced paradigm still provides a meaningful comparison with our approach. All other parameters are set to the default values reported in the original paper.

\subsection{LLM Evaluation Settings}
\label{sec:LLM Evaluation Settings}

Specifically, both the generated summary and the reference summary are first segmented into sentences, with each sentence treated as an individual event. For each generated sentence, we iterate over all reference sentences and employ an LLM to determine whether a semantic alignment exists. Precision measures the correctness of the generated summary, Recall reflects their coverage of reference summary, and F1 provides a balanced overall assessment. For the judging models, we adopt DeepSeek-v3~\cite{liu2024deepseek} (671B parameters, strong semantic capability) as the primary evaluator, and use Qwen3-32B~\cite{yang2025qwen3} as an auxiliary evaluator to enhance robustness and reliability. 

\begin{figure*}[!t]
    \centering
    \includegraphics[width=\textwidth]{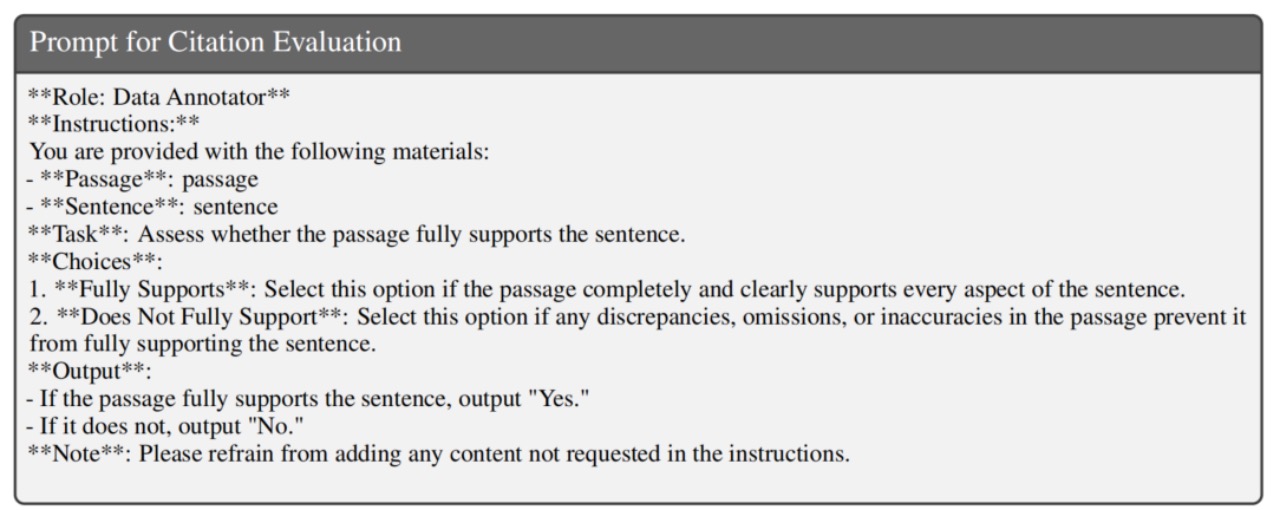}
    \caption{Prompt for LLM-based evaluation}
    \label{pic:LLM-prompt}
\end{figure*}

The prompt used in this work follows the template proposed in~\cite{sun2024towards}, as shown in Figure~\ref{pic:LLM-prompt}. In the original study, this prompt was designed to determine whether one sentence semantically supports another. Here, we extend its use to evaluate whether two sentences refer to the same event, which aligns well with the event-level semantic matching requirement in QFES. Therefore, we adopt this template directly for LLM-based evaluation.

\section{Detailed QFESum Settings}
\label{sec:Detailed QFESum Statics}

The Figure~\ref{Fig:QueryGenerationPrompt} shows the $\textit{query generation}$ prompt $\mathsf{P}_{\text{pg}}$ and an example. The Figure~\ref{fig:instruction generation prompt} shows the $\textit{instruction generation}$ prompt $\mathsf{P}_{\text{ig}}$. The Figure~\ref{fig:prompt_relevance_judgment} shows the $\textit{relevance judgement}$ prompt $\mathsf{P}_{\text{rj}}$. The Figure~\ref{pic:prompt_annatation} shows an example of the query "Treasury Department response and intervention" relevance judgement prompt $\mathsf{P}_{\text{rj}}$ that ChatGPT-4o generated to guide Qwen2.5-7B to annotate sentence-level relevance.

\definecolor{AccentRed}{RGB}{200,40,40}

\begin{figure}[t]
\vspace*{-0.6\baselineskip}
\centering

\begingroup
\setlength{\parindent}{0pt}
\setlength{\parskip}{0pt}

\fontsize{7.2pt}{8.4pt}\selectfont

\begin{tcolorbox}[
  enhanced,
  sharp corners,
  width=0.97\linewidth,
  colback=white,
  colframe=black!70,
  boxrule=0.4pt,
  arc=0.5pt, outer arc=0.5pt,
  boxsep=0pt,
  left=0.8pt,right=0.8pt,top=0.8pt,bottom=0.8pt,
  before skip=0pt, after skip=0pt,
  title={Prompt for Query Generation},
  colbacktitle=black!8,
  coltitle=black,
  fonttitle=\bfseries\fontsize{7.2pt}{8.4pt}\selectfont,
]

\begin{tcolorbox}[
  enhanced,
  sharp corners,
  boxrule=0pt,
  colback=blue!3,
  boxsep=0pt,
  left=0.8pt,right=0.8pt,top=0.6pt,bottom=0.6pt,
  before skip=0pt, after skip=0pt,
]
\raggedright
\textbf{Instructions:}\par
You are given a set of topic keywords extracted from a semantic cluster.
\textcolor{AccentRed}{Your task is to integrate these keywords into a concise, fluent,
and human-readable query phrase} that summarizes the underlying
semantic theme. The output should be suitable for use as a
query in a query-focused summarization task.
\end{tcolorbox}

\begin{tcolorbox}[
  enhanced,
  sharp corners,
  boxrule=0pt,
  colback=green!3,
  boxsep=0pt,
  left=0.8pt,right=0.8pt,top=0.6pt,bottom=0.6pt,
  before skip=0.6pt, after skip=0pt,
]
\raggedright
\textbf{Input Keywords:} \{\textit{\textcolor{AccentRed}{``revolution,'' ``revolutions,'' ``youth,'' ``revolutionaries,'' ``regimes,'' ``dictators''}}\}\par
\textbf{Output Query Phrase:} \{\textit{\textcolor{AccentRed}{``Youth-led revolutionary movements''}}\}
\end{tcolorbox}

\end{tcolorbox}
\endgroup
\vspace{6pt}
\caption{Prompt used Query Generation}
\label{Fig:QueryGenerationPrompt}
\end{figure}

\definecolor{AccentRed}{RGB}{200,40,40}

\begin{figure}[H]
\vspace*{-0.6\baselineskip}
\centering

\begingroup
\setlength{\parindent}{0pt}
\setlength{\parskip}{0pt}

\fontsize{7.2pt}{8.4pt}\selectfont

\begin{tcolorbox}[
  enhanced,
  sharp corners,
  width=0.97\linewidth,
  colback=white,
  colframe=black!70,
  boxrule=0.4pt,
  arc=0.5pt, outer arc=0.5pt,
  boxsep=0pt,
  left=0.8pt,right=0.8pt,top=0.8pt,bottom=0.8pt,
  before skip=0pt, after skip=0pt,
  title={Prompt for Instruction Generation},
  colbacktitle=black!8,
  coltitle=black,
  fonttitle=\bfseries\fontsize{7.2pt}{8.4pt}\selectfont,
]

\begin{tcolorbox}[
  enhanced,
  sharp corners,
  boxrule=0pt,
  colback=blue!3,
  boxsep=0pt,
  left=0.8pt,right=0.8pt,top=0.6pt,bottom=0.6pt,
  before skip=0pt, after skip=0pt,
]
\raggedright
\textbf{Instructions:}\par
\textcolor{AccentRed}{You are given a query and a reference summary focused on this query.}
All sentences in the summary are manually selected as relevant to the query.
Your task is to carefully read the reference summary and, in conjunction with the query,
\textcolor{AccentRed}{produce a prompt that helps a large language model judge whether a given sentence is relevant to the query.}
The prompt should include (i) a detailed semantic clarification of the query and (ii) illustrative positive and negative example sentences.
\end{tcolorbox}

\begin{tcolorbox}[
  enhanced,
  sharp corners,
  boxrule=0pt,
  colback=green!3,
  boxsep=0pt,
  left=0.8pt,right=0.8pt,top=0.6pt,bottom=0.6pt,
  before skip=0.6pt, after skip=0pt,
]
\raggedright
\textbf{Input Query:} \{\textit{Query}\}\par
\textbf{Input QFS Summary:} \{\textit{QFS Summary}\}\par
\textbf{Output Annotation Prompt:} \{\textit{Annotation Prompt}\}
\end{tcolorbox}

\end{tcolorbox}
\endgroup
\vspace{6pt}
\caption{Prompt for Instruction Generation}
\label{fig:instruction generation prompt}
\end{figure}

\definecolor{AccentRed}{RGB}{200,40,40}

\begin{figure}[H]
\vspace*{-0.6\baselineskip} 
\centering

\begingroup
\setlength{\parindent}{0pt}
\setlength{\parskip}{0pt}

\fontsize{7.2pt}{8.4pt}\selectfont  

\begin{tcolorbox}[
  enhanced,
  sharp corners,
  width=0.97\linewidth,
  colback=white,
  colframe=black!70,
  boxrule=0.4pt,          
  arc=0.5pt, outer arc=0.5pt,
  boxsep=0pt,             
  left=0.8pt,right=0.8pt,top=0.8pt,bottom=0.8pt,
  before skip=0pt, after skip=0pt,
  title={Prompt for Relevance Judgment},
  colbacktitle=black!8,
  coltitle=black,
  fonttitle=\bfseries\fontsize{7.2pt}{8.4pt}\selectfont,
]

\begin{tcolorbox}[
  enhanced,
  sharp corners,
  boxrule=0pt,
  colback=blue!3,
  boxsep=0pt,
  left=0.8pt,right=0.8pt,top=0.6pt,bottom=0.6pt,
  before skip=0pt, after skip=0pt,
]
\textbf{Instructions:}\par
\textcolor{AccentRed}{Please determine whether the following sentence is strictly relevant to the query: ``{query}''.}
Only answer ``True'' if the sentence explicitly discusses issues related to the query. Otherwise, respond ``False''. \par
Sentence: ``{sent}''\par
Answer with a single word: ``True'' or ``False''.
\end{tcolorbox}

\begin{tcolorbox}[
  enhanced,
  sharp corners,
  boxrule=0pt,
  colback=green!3,
  boxsep=0pt,
  left=0.8pt,right=0.8pt,top=0.6pt,bottom=0.6pt,
  before skip=0.6pt, after skip=0pt, 
]
\textbf{Input Query:} \{\textit{Query}\}\par
\textbf{Input Sentence:} \{\textit{Sentence}\}\par
\textbf{Output Labels:} \{\textit{True or False}\}
\end{tcolorbox}

\end{tcolorbox}
\endgroup
\vspace{6pt}
\caption{Prompt for Relevance Judgment}
\label{fig:prompt_relevance_judgment}
\end{figure}

\begin{table}[t]
\centering
\footnotesize
\setlength{\tabcolsep}{5pt}
\renewcommand{\arraystretch}{1.02}
\begin{tabular}{lccc}
\toprule
\textbf{Corpus} &
\textbf{min\_dist} &
\textbf{min\_cluster\_size} &
\textbf{min\_samples} \\
\midrule
BPoil          & 0.05  & 14 & 13 \\
finan          & 0.005 & 3  & 2 \\
iraq           & 0.05  & 10 & 4 \\
syria\_t17     & 0.05  & 15 & 15 \\
egypt          & 0.05  & 15 & 15 \\
libya          & 0.05  & 19 & 19 \\
syria\_crisis  & 0.05  & 19 & 23 \\
yemen          & 0.05  & 19 & 19 \\
\bottomrule
\end{tabular}
\caption{
Parameter settings for each corpus, including min\_dist, min\_cluster\_size, and min\_samples.
}
\label{tab:shc_args_settings}
\end{table}

\vspace{-4pt} 

\begin{table*}[t]
\centering
\footnotesize
\resizebox{\textwidth}{!}{
\begin{tabular}{lcccccccc}
\toprule
\textbf{Index} &
\textbf{BPoil} &
\textbf{finan} &
\textbf{iraq} &
\textbf{syria\_t17} &
\textbf{egypt} &
\textbf{libya} &
\textbf{syria\_crisis} &
\textbf{yemen} \\
\midrule
\textbf{gold\_documents\_num} & 222.57 & 75.19 & 82.07 & 286.12 & 786.70 & 599.29 & 1146.22 & 899.38 \\
\textbf{gold\_documents\_event\_num} & 875.81 & 275.48 & 277.50 & 1280.25 & 2176.60 & 1365.14 & 2959.11 & 1884.38 \\
\textbf{reference\_summary\_event\_num} & 21.43 & 28.43 & 16.00 & 30.12 & 18.40 & 17.36 & 21.62 & 18.88 \\
\textbf{generated\_summary\_event\_num} & 22.79 & 41.62 & 11.14 & 20.62 & 26.00 & 24.71 & 38.12 & 25.00 \\
\bottomrule
\end{tabular}
}
\caption{
Documents and summary event number across the eight corpus.
}
\label{tab:topics_summary_lengths}
\end{table*}

\begin{figure*}[t]
\centering
\begin{minipage}{\textwidth}

\begin{tcolorbox}[promptbox={Prompt for Relevance Annotation}, width=\textwidth]

\scriptsize

\setlength{\parskip}{2pt}
\setlength{\itemsep}{1pt}
\setlength{\topsep}{1pt}
\setlength{\partopsep}{0pt}


You are given a query and a sentence. Your task is to classify whether the sentence is relevant to the query.
The following are some guidance about the task.

\medskip
\textbf{Query:}
``Treasury Department response and intervention''

\medskip
This query refers to actions, policies, announcements, or decisions made by the query \texttt{"U.S. Department of the Treasury"} in response to the 2008 financial crisis. Relevant sentences include:

\begin{itemize}
    \item Statements or public addresses by \textbf{Treasury Secretary Henry Paulson}
    \item Treasury-led programs like the \textbf{\$700 billion financial rescue (TARP)} and other bailout efforts
    \item Decisions to inject capital into banks, buy mortgage-backed securities, or stabilize money-market funds
    \item Appointments or hiring of officials (e.g., Neel Kashkari) or legal/financial advisors to oversee intervention programs
    \item Testimony or defenses of Treasury actions before Congress
    \item Joint actions with the Federal Reserve or FDIC where Treasury plays a lead role
    \item Any Treasury announcements regarding guarantees, investments, or expanded rescue efforts
\end{itemize}

\medskip
\textbf{Here are examples of relevant sentences:}

\begin{enumerate}
    \item ``Henry Paulson says the government will start buying securities backed by mortgages -- initially, up to \$5 billion worth.''
    \item ``Treasury Secretary Henry Paulson urges patience as global markets continue to slide.''
    \item ``Treasury Secretary Paulson lays out steps to shore up the nation's banking system by having the government buy shares in leading banks.''
    \item ``Treasury Dept. says it will announce that a group of large regional banks have agreed to accept investments from the government, joining nine of the largest American banks who were forced to accept \$125 billion in funding last week.''
    \item ``The Treasury and FDIC say they are crafting a plan under which the government would guarantee the mortgages of as many as 3 million homeowners now struggling to avoid foreclosure.''
\end{enumerate}

\medskip
\textbf{Here are examples of sentences that are irrelevant:}

\begin{enumerate}
    \item ``The government seized control of the insurance giant American International Group to preserve a crucial bulwark of the global financial system.''
    \item ``The Federal Reserve was forced to ask the Treasury yesterday to borrow some extra money to replenish its coffers.''
    \item ``Democrats want to create a new agency that would use money borrowed by the Treasury to recapitalize troubled financial institutions by buying some of their unwanted loans and securities at discounted prices.''
    \item ``House Speaker Nancy Pelosi dispatched Rep. Barney Frank to determine whether Federal Reserve Chairman Ben S. Bernanke should retain authority to unilaterally bail out failing firms.''
    \item ``Bush has left direct management of the crisis to them and other advisers, and has limited his public remarks on the economy.''
\end{enumerate}

\medskip
Sentences are \textbf{not relevant} if they only describe the Federal Reserve or the government's response to the economic crisis, or if they only describe the requests and demands of other institutions or departments to the Treasury Department, without describing the measures and actions taken by the Treasury Department.

Sentences \textbf{are relevant} only if they directly refer to the measures and actions taken by the Treasury Department.

\medskip
Now determine whether the following sentence is strictly relevant to the query.

\medskip
\noindent\textbf{Query:} ``\verb|{query}|''

\medskip
\noindent\textbf{Sentence:} ``\verb|{sentence}|''

\medskip
\textbf{Respond with only:}

\begin{center}
\texttt{true} or \texttt{false}
\end{center}

Do not include any explanation or additional text.

\end{tcolorbox}

\end{minipage}
\caption{Prompt for Relevance Annotation}
\label{pic:prompt_annatation}
\end{figure*}

\end{document}